\pdfoutput=1

\documentclass[11pt]{article}

\usepackage{EMNLP2023}

\usepackage{times}
\usepackage{latexsym}
\usepackage{booktabs}
\usepackage{amssymb}
\usepackage{amsmath}
\usepackage{graphicx}

\usepackage[T1]{fontenc}

\usepackage[utf8]{inputenc}

\usepackage{microtype}

\usepackage{inconsolata}

\title{BabyStories: Can Reinforcement Learning Teach Baby Language Models to Write Better Stories?}

\author{Xingmeng Zhao, Tongnian Wang, Sheri Osborn, \and Anthony Rios  \\
  Department of Information Systems and Cyber Security \\
  The University of Texas at San Antonio \\
  \texttt{\{xingmeng.zhao, tongnian.wang, sheri.osborn, anthony.rios\}@utsa.edu} \\}

\begin{document}
\maketitle

\begin{abstract}

Language models have seen significant growth in the size of their corpus, leading to notable performance improvements. Yet, there has been limited progress in developing models that handle smaller, more human-like datasets. As part of the BabyLM shared task, this study explores the impact of reinforcement learning from human feedback (RLHF) on language models pretrained from scratch with a limited training corpus. Comparing two GPT-2 variants, the larger model performs better in storytelling tasks after RLHF fine-tuning. These findings suggest that RLHF techniques may be more advantageous for larger models due to their higher learning and adaptation capacity, though more experiments are needed to confirm this finding. These insights highlight the potential benefits of RLHF fine-tuning for language models within limited data, enhancing their ability to maintain narrative focus and coherence while adhering better to initial instructions in storytelling tasks. The code for this work is publicly at \url{https://github.com/Zephyr1022/BabyStories-UTSA}.

\end{abstract}

\section{Introduction}

The recent growth in the size of large language models (LLMs) has enhanced natural language processing capabilities, from information extraction~\cite{agrawal2022large} to language generation~\cite{stiennon2020learning}. However, the majority of research has been concentrated on environments with high computational power and a large number of parameters, leaving the emergence of these capabilities largely uninvestigated in low data and low resource settings~\cite{brown2020language,fedus2022switch}. Although some studies have looked into the relationship between model size, training volume, and performance for LLMs, they have primarily focused on scaling laws in high-compute settings~\cite{hoffmann2022training}. Investigations into the effects of pretraining at a smaller scale have been limited~\cite{huebner2021babyberta,deshpande2023honey}. Therefore, it would be interesting to explore strategies that maximize the efficiency of pretraining, especially considering the constraints of limited data availability. 

Storytelling is a fundamental human activity used to share information, impart lessons, and keep loved ones informed about our daily lives~\cite{bietti2019storytelling}. Teachers leverage children's love for stories and their desire to tell them, using storytelling to promote cognitive and literacy development. Storytelling is a critical bridge between the oral language skills of early childhood and the more mature language skills associated with reading and writing. The recent BabyLM shared task aims to address these challenges~\cite{warstadt-et-al-2023-babylm}. Hence, we report our submission to the shared task in this paper. Specifically, our study aims to understand whether we can pretrain a language model from scratch on the same amount of linguistic data available to a child, modeling a smaller, reduced-vocabulary language. We are interested in assessing a particular model's effectiveness and potential for enhancement. Specifically, we investigate whether the model can demonstrate high performance and if its performance can be further improved using reinforcement learning techniques from human feedback (RLHF)~\cite{fernandes2023bridging}. This process is analogous to how teachers instruct children in storytelling, providing feedback to encourage them to develop more coherent and reasonable narratives. Implementing RLHF has shown promising results in aligning foundation models with human preferences. By using RLHF, models can undergo subtle yet significant improvements, such as refining tone~\cite{liu2023perspectives}, reducing biases and toxic elements~\cite{bai2022training}, and enabling domain-specific content generation~\cite{bang2023multitask}. The primary goal of this research is to explore whether the small pretrained model, with its limited data size, can also benefit from RLHF, thus potentially improving its overall performance.

The performance of small language models (SLMs) trained on large datasets has been observed to be poor, generating incoherent and repetitive text. Training large language models on limited data can lead to overfitting, making smaller models a potential solution to prevent overfitting~\cite{warstadt-etal-2020-learning}. Inspired by how humans acquire language and the BabyLM shared task, we explore downsizing the language used in models to observe the effects of pretraining. The main questions are whether small language models can generate coherent English text and if this ability is limited to larger, more complex models.
It is also questioned whether the limited capacity of small models to memorize linguistic features—such as syntax, semantics, morphology, and phonology—leads to less creative outputs compared to larger models. For example, linguistic features are crucial for understanding and generating text, with a broader grasp potentially enabling more creative language use. Larger models, with their increased capacity, might capture a wider range of these features, possibly leading to more creative and nuanced language outputs. Conversely, small models might only learn basic or frequent linguistic patterns, potentially limiting their creative language generation capabilities.
Previous research indicates that models can learn linguistic features with limited pretraining data but need more data to prioritize linguistic generalizations over superficial ones~\cite{warstadt-etal-2020-learning}. Some models fail to effectively use the linguistic features they learn during fine-tuning for natural language understanding tasks. The study aims to investigate whether GPT-2 models of varying sizes can acquire specific language patterns when fine-tuned with reinforcement learning and human feedback, aiming to enhance the model's storytelling abilities.


In summary, in this paper, we pretrain GPT-2-base model with a parameter of 125M from scratch and compare it with the larger GPT2-Large model, which has a parameter of 774M, making it approximately six times larger. Both models are trained using a limited dataset provided from the BabyLM Challenge, which consists of approximately 100M words~\cite{warstadt-et-al-2023-babylm}.
The dataset encompasses various sources, including child-directed speech, transcribed speech from multiple sources, children's books, and Wikipedia. Subsequently, we use the RLHF technique to fine-tune both models and evaluate their ability to acquire new linguistic features through human feedback and also perform human evaluation on generated stories. 


\section{Related Work}

Research has shown that smaller models tend to underperform when trained on large datasets, making the study of model downscaling a non-trivial~\cite{turc2019well}. Previous investigations into smaller models have primarily centered around distillation processes~\cite{sanh2019distilbert}, with the aim of maximizing performance while reducing the number of parameters involved. \citet{huebner2021babyberta} is one of the most relevant papers to our work, where they found that a small language model trained on child-directed speech can yield results comparable to larger language models when used in specific probing tasks. And another study, \citet{deshpande2023honey} trained several models to explore scaling in low-compute environments, assessing their performance on a modified version of GLUE.

Our research, however, is driven by a desire to understand if small pretrained models can benefit from Reinforcement Learning from Human Feedback (RLHF), potentially improving their overall performance despite their limited data size. Two previous studies have a direct relation to this work: the first employed human ranking feedback to train summarization models using reinforcement learning (RL)~\cite{stiennon2020learning}, and the second used stories to generate a value-aligned reward signal for RL agents, aimed at mitigating hallucination behavior~\cite{riedl2016using}.

\section{Data}

In this section, we describe the pertaining data used for the language models and the data used for the reinforcement model.

\subsection{Pretraining Data}
We pretrain GPT-2 models using the dataset from the STRICT track in the BabyLM Challenge~\cite{warstadt-et-al-2023-babylm}, which includes various types of corpora, both spoken-based and written-based. Examples of the spoken-based corpus include CHILDES \cite{macwhinney2000childes}, the British National Corpus (BNC) dialogue section, OpenSubtitles \cite{lison2016opensubtitles2016}, the QCRI Educational Domain Corpus \cite{abdelali2014amara}, and the Switchboard Dialog Act Corpus \cite{stolcke2000dialogue}. The written-based corpus includes the Children’s Book Test \cite{hill2016goldilocks}, the Children’s Stories Text Corpus, the Standardized Project Gutenberg Corpus \cite{gerlach2020standardized}, Wikipedia, and Simple Wikipedia. For example, the Children's Book Story and Wikipedia corpora stand in contrast to dialogue or subtitle-based corpora, which mostly consist of transcribed speech, the primary language input for children. Wikipedia, in particular, is a compilation of written language rather than spoken dialogues. Most of its articles are composed by professionals who possess subject-matter expertise and adhere to rigorous standards of grammatical correctness. These corpora contain a variety of sources with approximately 100 million words, corresponding to the linguistic competence expected at the onset of adolescence (around 13 years old).

\subsection{Reward Model Data}

In this paper, we construct a reward model dataset for reinforcement learning by selecting 100 sentences from the STRICT track of the Babylm Challenge dataset. These sentences, serving as prompts, are derived from two subsets in the Babylm dataset: the Standardized Project Gutenberg and the Simple Wikipedia corpus development sets, with a prerequisite that each sentence includes characters and plots. These prompts are then used to generate two short stories each from the GPT-2 Base and GPT-2 Large models, beginning with the prefix ``write me a story starting with''. To enhance story diversity, we set a maximum length of 128 tokens and enforce a minimum of 10 new tokens in the generated stories. The generation code incorporates a beam size of 7 to optimize the story quality by exploring various potential continuations. 

The purpose of collecting feedback is to align the model’s behavior with some goal behavior. For example, we aim for the model to generate stories consistent with the background plot, coherent, non-repetitive, devoid of nonsensical sentences, and maintain a clear topic or logical structure. Rating the quality of a story accurately presents challenges due to its potentially subjective nature and the varying expectations of readers regarding emotional connection and engagement. Rather than directly estimating a generated story quality through scale-based annotation, we treat it as a latent variable to be inferred from relative comparisons. Following prior work in NLP on annotating social aspects of language~\cite{pei2020quantifying}, we adopt a method similar to Best-Worst Scaling (BWS)~\cite{louviere2015best,kiritchenko2016capturing} to generate comparison data on people's preferences. Intuitively, it is easier for annotators to identify the best and worst stories from a set of stories than it is for them to provide numerical assessments. The process involves asking two student annotators to choose from sets of stories, identifying the best (most preferred) and worst (least preferred) stories in each choice set. We provide four stories for the annotators to choose from. This method provides more information per choice set than traditional preference methods and enables a more precise ranking of items in terms of preference. For instance, if we have stories A, B, C, and D, and A is ranked as the best while D is ranked as the worst, then we create the following pairs: A $>$ B, A $>$ C, A $>$ D, B $>$ D, and C $>$ D, resulting in a total of 500 additional pairs for reward model training from 100 best-worst annotations. A $>$ B means that the model should learn to provide a higher score to A because it was ranked higher than B. This is inferred because A was marked as the best story.

\subsubsection{Agreement for Reward Model Data Annotation}
Krippendorff's alpha, introduced by~\citet{krippendorff1970estimating}, is a statistical measure commonly used for assessing the level of agreement between two or more annotators across various categories. Its advantage lies in its versatility, as it can be applied to not only nominal data but any measurement scale, such as Best-Worst Scaling. 


In our case, two graduate student annotators were designated to annotate human feedback data., which yielded a Krippendorff's alpha agreement score of .4657. To address disagreements, the two annotators discuss each story example together. They reconcile differences through discussion and unanimously select the best and worst stories based on the given story prompt.

\section{Method}

This section discusses pretraining data, the development of the data tokenizer, language model configuration, the objective of pretraining from scratch, and the process of fine-tuning using reinforcement learning with human feedback.

\subsection{Tokenizer}
 
Our model uses a sub-word vocabulary built with Byte-Pair Encoding (BPE)~\cite{sennrich2016edinburgh}, an approach initially developed for text compression. Later, this technique was applied by OpenAI for tokenization during the pretraining stage of the GPT model~\cite{radford2019language}. Rather than maintaining the original vocabulary size of 50,257 used in the GPT-2 model, we developed a custom tokenizer based on a vocabulary size of 32,001. This custom tokenizer is trained on the collective set of all training corpora from STRICT track in the BabyLM Challenge, applying the ByteLevelBPETokenizer from the Hugging Face Tokenizers library\footnote{\url{https://github.com/huggingface/tokenizers}}.

Prior research informed our decision to significantly reduce the vocabulary size. Studies suggest a vocabulary size of about 32,000 tokens is a good balance for a single-language model~\cite{kudo-2018-subword}. This size carefully balances the model's proficiency in handling less common words while preserving its computational efficiency. 


\subsection{Model Architecture and Configuration}


Models we pretrained in our experiments using the default configuration setting of GPT-2~\cite{radford2019language}. In these settings, we employed a context length of 1042 tokens and set the maximum training epoch limit to 15. The restriction to 15 epochs was primarily due to constraints on training time and GPU resources. We conducted the training of the GPT-2 Base model on an NVIDIA GeForce GTX 1080 Ti, while the GPT-2 Large model was trained on an NVIDIA RTX A6000 GPU. The training time for the base model was approximately 72 hours, while it extended to around 216 hours for the large model. To train, we used the Lion optimizer~\cite{chen2023symbolic}, configured with a learning rate of 1e-5 and a weight decay of 1e-2. We also integrated Triton, a GPU programming language detailed by~\cite{tillet2019triton}, to optimize hardware performance and implemented mixed-precision computations using the 'bfloat16' format for efficient resource utilization~\cite{wang2019bfloat16}.

For model selection, we chose the best model across all epochs based on the average score on two datasets: the Question-answering Natural Language Inference (QNLI)~\cite{demszky2018transforming} and the SST-2 Binary Classification Benchmark~\cite{socher2013recursive}. We evaluated the models' performances on these benchmarks using the F1 score. Additionally, the perplexity scores on the validation dataset for our models were recorded as 24.10 for the GPT-2 Base model and 22.73 for the GPT-2 Large model.


\subsection{Reward Model}

The reward model (RM) is designed to capture human preferences, and ideally, we could fine-tune it using Reinforcement Learning and human annotations for every output returned by the language model. However, due to practical constraints like workload and time limitations, it is not feasible for humans to provide enough feedback for each optimization iteration. As an alternative, a more effective approach is to train a reward model that simulates the evaluation process carried out by humans. This RM will evaluate any text and assign a scalar reward value to the sentences, where higher values indicate high-quality samples. Following \citet{stiennon2020learning}, training reward models often involve using a paired comparison dataset between two responses generated for the same input.

To train our reward models, We initialize the weights of the reward model by leveraging a pre-trained GPT-2 Large model as described above, then we add a randomly initialized linear head that outputs a scalar value to form the reward model $r_{\theta}(x, y)$. We train this model to predict which generated story $y \in \{y_0, y_1\}$, where $y_0$ is the chosen (good) response to the prompt as labeled by our annotators and $y_1$ is the rejected (bad) response. In practice, this is where our annotators ranked $y_0 > y_1$. The model is trained using the loss function 
\begin{align*}
        loss(r_{\theta}) = -E_{(x,y_0,y_1,i)\sim D}\Big[\log(\sigma(r_{\theta}(x^i,y^i_0) \\ -\, r_{\theta}(x^i,y^i_{1}))\Big]
\end{align*}
where $\sigma$ is the sigmoid function and $D$ is the set of all training triplets in our dataset, i denotes the index of a specific data point in the dataset D. Intuitively, the model learns to give a larger score to the prompts with a higher rank. We have configured the reward model to run for a maximum of 10 epochs, with a set learning rate of 1e-5.


\subsubsection{Proximal Policy Optimization}

After we train the reward model, we treat the logit output of the reward model as a reward that we optimize policy model outputs using reinforcement learning, specifically with the Proximal Policy Optimization (PPO) algorithm~\cite{schulman2017proximal}. During the RL fine-tuning with PPO phase, we use the learned reward function to provide feedback to the language model. In particular, we formulate the following optimization problem
\begin{align*}
\max_{\pi^{RL}(y|x)} E_{x \sim D, y \sim \pi^{RL}(y|x)} \Big[r(x,y)\Big] \\ -\, \beta D_{KL}\log\left[\frac{\pi^{RL}(y|x)}{\pi^{SFT}(y|x)}\right]
\end{align*}
where $r(x, y)$ is the reward model's output, $\beta$ is a hyper-parameter controlling the deviation from the initial policy. Our optimization focuses on the policy $\pi^{RL}(y|x)$ using Proximal Policy Optimization (PPO), with initialization based on the pretrained language model policy $\pi^{SFT}(y|x)$~\cite{stiennon2020learning,rafailov2023direct}. 

To encourage exploration and prevent the policy from getting stuck in a single mode, the optimization uses the Kullback-Leibler (KL) divergence term. This term also discourages the policy from generating outputs that differ significantly from those seen by the reward model during training, thereby maintaining coherence in the generated text. Without this penalty, the optimization might generate gibberish text that tricks the reward model into providing a high reward. In our implementation, we used the trlX library with its default settings\footnote{\url{https://github.com/CarperAI/trlx}}. The algorithm was executed with a maximum of 5 epochs and a sequence length of 512, and the run spanned around 208 hours. In our approach, we used the default hyperparameter provided by the trlX library, which employs Ray Tune for hyperparameter tuning. This choice was primarily driven by the significant time and GPU resource constraints associated with training the PPO model, making it a pragmatic decision to leverage the pre-configured settings of trlX. Although we experimented with random modifications to some hyperparameters, the outcomes were less satisfactory as compared to the default settings of trlX. Hence, the decision to restrict the training to 5 epochs was in alignment with these considerations, ensuring a balance between computational feasibility and the pursuit of meaningful reward training.


\subsection{Evaluation Metrics and Datasets}

To assess the performance of our models, we employed various automated evaluation metrics used in the BabyLM shared task and our own human evaluation. The BabyLM shared task had two major sets of evaluations: zero-shot evaluation and fine-tuned evaluation. 
We describe each evaluation task below.

\begin{table*}[t]
\resizebox{\textwidth}{!}{%
\begin{tabular}{@{}lrrrrrrrrrrrrrrrrrr@{}}
\toprule
\multicolumn{1}{c}{\textbf{Model}} &
  \multicolumn{1}{c}{\textbf{AA}} &
  \multicolumn{1}{c}{\textbf{AS}} &
  \multicolumn{1}{c}{\textbf{BD}} &
  \multicolumn{1}{c}{\textbf{CR}} &
  \multicolumn{1}{c}{\textbf{DNA}} &
  \multicolumn{1}{c}{\textbf{E}} &
  \multicolumn{1}{c}{\textbf{FG}} &
  \multicolumn{1}{c}{\textbf{IF}} &
  \multicolumn{1}{c}{\textbf{IE}} &
  \multicolumn{1}{c}{\textbf{NL}} &
  \multicolumn{1}{c}{\textbf{Q}} &
  \multicolumn{1}{c}{\textbf{SV}} &
  \multicolumn{1}{l}{\textbf{H}} &
  \multicolumn{1}{l}{\textbf{QACe}} &
  \multicolumn{1}{l}{\textbf{QACt}} &
  \multicolumn{1}{l}{\textbf{SAI}} &
  \multicolumn{1}{l}{\textbf{TT}} &
  \multicolumn{1}{l}{\textbf{AVG}} \\ \midrule 
 \multicolumn{19}{c}{\textbf{Baselines}} \\ \midrule
OPT-125m         & 94.9 & 73.8 & 73.8 & \textbf{72.2} & \textbf{93.1} & 80.5 & 73.6 & 80.8 & 57.8 & 51.6 & 74.5 & 77.3 & 46.3 & \textbf{76.5} & 47.9 & 85.3 & \textbf{82.9} & 73.1 \\
RoBERTa-base     & 89.5 & 71.3 & 71.0 & 67.1 & \textbf{93.1} & \textbf{83.8} & 68.0 & 89.6 & 54.5 & 66.3 & 70.3 & 76.2 & 50.8 & 34.4 & 34.5 & 45.6 & 46.8 & 65.5\\
T5-base          & 66.7 & 61.2 & 59.4 & 59.8 & 53.8 & 49.1 & 70.0 & 75.5 & 43.6 & 45.6 & 34.2 & 53.2 & \textbf{51.1} & 45.3 & 25.5 & 69.2 & 48.9 & 53.7 \\ \midrule \midrule
\multicolumn{19}{c}{\textbf{Ours}} \\ \midrule
GPT2-Base  & 95.4 & 75.5 & 74.0 & 67.0 & 90.8 & 77.7 & 70.0 & \textbf{87.7} & 53.6 & 57.6 & \textbf{79.0} & 75.8 & \textbf{50.2} & 60.9 & 41.8 & 85.0 & 67.9 & 71.2 \\
GPT2-Base-PPO  & 95.5 & 75.4 & 73.6 & 67.0 & 90.8 & 78.3 & 70.2 & 86.7 & 54.4 & 58.0 & 77.7 & 75.2 & 49.9 & 59.4 & 40.0 & \textbf{85.7} & 68.2 & 70.9 \\ \midrule
GPT2-Large & 96.9 & 78.7 & \textbf{74.1} & \textbf{71.0} & 92.0 & 79.0 & 73.8 & 87.2 & \textbf{60.8} & \textbf{60.9} & 75.9 & \textbf{81.1} & 49.2 & \textbf{71.9} & 49.7 & 79.8 & \textbf{73.6} & \textbf{73.9} \\
GPT2-Large-PPO & \textbf{97.0} & \textbf{78.8} & \textbf{74.1} & \textbf{71.0} & \textbf{92.1} & \textbf{79.3} & \textbf{73.7} & 87.1 & 60.7 & 60.8 & 75.9 & \textbf{81.1} & 49.4 & \textbf{71.9} & \textbf{50.3} & 79.6 & 73.2 & \textbf{73.9} \\ \bottomrule
\end{tabular}%
}
\caption{Performance on BLiMP benchmarks. Evaluation tasks map accordingly: Anaphor Agr.:AA, 	Agr. Structure: AS, Binding: BD, Control/Raising: CR, D-N Agr.: DNA, Ellipsis: E, Filler-Gap: FG, Irregular Forms: IF, Island Effects: IE, NPI Licensing: NL, Quantifiers: Q, S-V Agr.: SV, Hypernym: H, QA Congruence (easy): QAC(e), QA Congruence (tricky): QAC(t), Subj.-Aux. Inversion: SAI, Turn Taking: TT. The overall largest scores are in bold. }
\label{tab:blimp}
\vspace{-1em}
\end{table*}

\paragraph{Zero-shot Evaluation.} BLiMP, introduced by \citet{warstadt2020blimp}, is a series of zero-shot tasks included in the evaluation. BLiMP assesses the ability of language models to handle category membership, provide congruent answers to specific types of questions, and recognize grammatical questions. It serves as a behavioral probe, containing pairs of test sentences that isolate particular phenomena in syntax and morphology, such as island effects and determiner-noun agreement. Essentially, BLiMP is a challenge set designed to evaluate the linguistic knowledge of language models, focusing on major grammatical phenomena in English. The BLiMP Supplement benchmark consists of BLiMP-style minimal pairs that specifically focus on aspects not covered by BLiMP. These additional aspects include discourse-level acceptability across multiple speakers and question formation.

\paragraph{Fine-tuned Evaluation.} Two datasets are used for the fine-tuned evaluation: SuperGLUE and the Mixed Signals Generalization Set (MSGS). SuperGLUE~\cite{wang2019superglue}, an advanced version of GLUE~\cite{wang2018glue}, is a benchmark for assessing progress in general-purpose language understanding technologies. It comprises a public leaderboard and a single-number performance metric for various tasks. These include CoLA, which evaluates the grammatical acceptability of English sentences; SST-2, which predicts the sentiment of movie review sentences; MRPC, which determines semantic equivalence between sentence pairs; QQP, another task focused on semantic equivalence; MNLI and MNLI-mm, which predict the relationship between a premise and a hypothesis sentence; QNLI, which matches a question to a paragraph containing the answer; RTE, which determines if a sentence entails a given hypothesis; BoolQ, which answers yes/no questions about a text passage; MultiRC, which identifies true and false answers given a context paragraph and a question; and WSC, a coreference resolution task. These tasks, designed to be challenging, represent a broad spectrum of language understanding capabilities, making SuperGLUE a robust tool for evaluating language models.

The MSGS dataset, introduced by ~\citet{warstadt2020learning}, is a diagnostic tool designed to evaluate the preferences of language models for either linguistic features, such as specific syntactic constructions, or surface features, like the presence of a word in a certain position. The primary objective of the MSGS tasks is to determine whether a pretrained model leans more toward linguistic or surface generalizations during the fine-tuning process. Fine-tuning on self-supervised linguistic tasks proves effective because it equips models with features beneficial for language understanding. Furthermore, pretrained models are not only capable of representing these linguistic features but also tend to use them preferentially during fine-tuning.

To maintain consistency and ensure fair comparisons, we adopted the default hyperparameter settings recommended by \citet{eval-harness}. Our only modification was adjusting the batch size to 32 due to GPU limitations. These evaluation procedures allowed us to thoroughly assess the models' capabilities and compare their performance across different tasks. Our experiments report the average scores of all performance metrics across tasks.

\begin{table*}[t]
\resizebox{\textwidth}{!}{%
\begin{tabular}{@{}lrrrrrrrrrrrr@{}}
\toprule
\textbf{Model} &
  \textbf{CoLA} &
  \textbf{SST-2} &
  \textbf{MRPC (F1)} &
  \textbf{QQP (F1)} &
  \textbf{MNLI} &
  \textbf{MNLI-mm} &
  \textbf{QNLI} &
  \textbf{RTE} &
  \textbf{BoolQ} &
  \textbf{MultiRC} &
  \textbf{WSC} &
  \textbf{AVG} \\ \midrule  
\multicolumn{13}{c}{\textbf{Baselines}} \\ \midrule
  Majority label & 69.5 & 50.2 & 82.0 & 53.1 & 35.7 & 35.7 & 35.4 & 53.1 & 50.5 & 59.9 & 53.2 & 52.6\\
OPT-125m       & 73.7 & 86.6 & 82.1 & 77.8 & 70.1 & 71.9 & 80.1 & \textbf{67.7} & 66.0 & 61.1 & 59.0 & 72.4\\
RoBERTa-base   & 75.9 & \textbf{88.6} & 80.5 & 78.5 & 68.7 & \textbf{78.0} & 82.3 & 51.5 & 59.9 & 61.3 & 61.4 & 71.5 \\
T5-base        & \textbf{76.3} & 88.0 & \textbf{85.9} & \textbf{79.7} & \textbf{71.5} & 74.0 & \textbf{83.1} & 60.6 & \textbf{69.0} & \textbf{62.4} & 60.2 & \textbf{73.7}\\ \midrule \midrule
\multicolumn{13}{c}{\textbf{Ours}} \\ \midrule
GPT2-Base      & 69.5 & 83.3 & 78.1 & \textbf{72.2} & 60.0 & 61.3 & 57.0 & 49.5 & 59.9 & 46.8 & 42.2 & 61.8 \\
GPT2-Base-PPO  & 69.5 & 81.3 & 82.0 & 67.3 & 60.9 & 61.7 & 61.4 & 45.5 & 59.9 & 46.8 & 39.8 & 61.5 \\ \midrule
GPT2-Large     & 69.5 & 82.7 & \textbf{83.0} & 32.4 & \textbf{61.4} & 62.2 & 54.4 & \textbf{58.6} & 66.8 & 46.8 & \textbf{61.5} & 61.7 \\
GPT2-Large-PPO & 69.5 & \textbf{84.3} & 82.3 & 66.7 & 59.5 & \textbf{64.0} & \textbf{79.6} & 53.5 & \textbf{67.4} & 46.8 & \textbf{61.5} & \textbf{66.8} \\ \bottomrule
\end{tabular}%
}
\caption{Performance on (Super)GLUE benchmarks. The task shortcuts correspond to the following datasets: Corpus of Linguistic Acceptability (CoLA), the Stanford Sentiment Treebank (SST-2), the Microsoft Research Paraphrase Corpus (MRPC), the Quora Question Pairs (QQP), the Multi-Genre Natural Language Inference (MNLI), the mismatched version of MNLI (MNLI-mm), the Question Natural Language Inference (QNLI), the Recognizing Textual Entailment (RTE), the BoolQ, the Multi-Sentence Reading Comprehension (MultiRC), and the Winograd Schema Challenge (WSC). The overall largest scores are in bold.}
\label{tab:my-glue}
\end{table*}

\begin{table*}[t]
\resizebox{\textwidth}{!}{%
\begin{tabular}{@{}lrrrrrrrrrrrr@{}}
\toprule
\multicolumn{1}{l}{\textbf{Model}} &
  \multicolumn{1}{r}{\textbf{CR\_C}} &
  \multicolumn{1}{r}{\textbf{LC\_C}} &
  \multicolumn{1}{r}{\textbf{MV\_C}} &
  \multicolumn{1}{r}{\textbf{RP\_C}} &
  \multicolumn{1}{r}{\textbf{SC\_C}} &
  \multicolumn{1}{r}{\textbf{CR\_LC}} &
  \multicolumn{1}{r}{\textbf{CR\_RTP}} &
  \multicolumn{1}{r}{\textbf{MV\_LC}} &
  \multicolumn{1}{r}{\textbf{MV\_RTP}} &
  \multicolumn{1}{r}{\textbf{SC\_LC}} &
  \multicolumn{1}{r}{\textbf{SC\_RP}} &
  \multicolumn{1}{r}{\textbf{AVG}} \\ \midrule
\multicolumn{13}{c}{\textbf{Baselines}} \\ \midrule
OPT-125m           & \textbf{97.2} & 82.6  & \textbf{100.0} & 99.8  & 88.1 & 75.3 & 67.1 & 66.3 & 66.8 & \textbf{84.8} & 62.0 & 80.9 \\
RoBERTa-base       & 93.0 & \textbf{100.0} & \textbf{100.0} & \textbf{100.0} & 89.0 & 68.3 & 66.8 & 66.6 & \textbf{80.2} & 67.4 & 67.4 & 81.7 \\
T5-base            & 95.1 & \textbf{100.0} & \textbf{100.0} & 99.8  & 88.7 & \textbf{76.7} & \textbf{69.4} & \textbf{67.0} & 67.7 & 72.7 & 68.0 & 82.3 \\ \midrule \midrule
\multicolumn{13}{c}{\textbf{Ours}} \\ \midrule
GPT2-Base          & \textbf{96.7} & \textbf{99.8}  & 99.7  & \textbf{100.0} & 95.5 & 68.2 & 68.3 & 66.6 & \textbf{67.0} & \textbf{74.6} & \textbf{76.5} & \textbf{83.0} \\
GPT2-Base-PPO      & 85.9 & \textbf{99.8}  & \textbf{99.9}  & 99.9  & 93.3 & \textbf{71.7} & 67.9 & 66.6 & \textbf{67.0} & 68.4 & 70.5 & 81.0 \\ \midrule
GPT2-Large         & 91.2 & 98.5  & \textbf{99.9}  & \textbf{100.0} & 94.0 & 67.3 & \textbf{68.5} & 66.6 & 66.8 & 71.9 & 69.4 & 81.3 \\
GPT2-Large-PPO & 93.6 & \textbf{99.8}  & 99.4  & \textbf{100.0} & \textbf{96.2} & 70.0 & 66.7 & 66.6 & 66.9 & 73.1 & 68.1 & 81.9 \\ \bottomrule
\end{tabular}%
}
\caption{Performance on MSGS benchmarks. The MSGS shortcuts correspond to the respective tasks as follows: CR\_RTP maps to control\_raising\_relative\_token\_position, CR\_LC maps to control\_raising\_lexical\_content\_the, SC\_RP maps to syntactic\_category\_relative\_position, SC\_LC maps to syntactic\_category\_lexical\_content\_the, MV\_RTP maps to main\_verb\_relative\_token\_position, MV\_LC maps to main\_verb\_lexical\_content\_the. The shortcuts RP\_C, LC\_C, SC\_C, CR\_C, and MV\_C correspond to the tasks relative\_position\_control, lexical\_content\_the\_control, syntactic\_category\_control, control\_raising\_control, and main\_verb\_control, respectively. The overall largest scores are in bold.}
\label{tab:my-msgs}
\end{table*}

\paragraph{Human Evaluation.} Inspired by the TinyStories~\cite{eldan2023tinystories}, we assess the four key story generation outcome metrics of grammar (how grammatically correct the story is), creativity (how original and inventive the story is), consistency with the story's beginning (how well the story adheres to the given prompts), and plot coherence (whether the plot of the story makes sense). We randomly selected 100 prompts from the ROCStories dataset~\cite{mostafazadeh-etal-2016-corpus}. Each prompt was composed of a story title and the first sentence. We fed these prompts to the model, and it generated short stories based on the given prompts. To assess the quality of the generated stories, we enlisted the help of a graduate student evaluator. The evaluator was presented with the story's beginning (title + first sentence) and the completed story generated by the model. They were then asked to rate the completed story on a scale of 1 to 10, considering aspects such as grammar, creativity, consistency with the story's beginning, and plot coherence. This human evaluation process provided valuable insights into the model's performance across these critical dimensions.


\section{Results}

In this section, we report the results of the automated BabyLM metrics and our human evaluation for story generation.

\paragraph{Performance on BLiMP benchmarks.} Shown in Table~\ref{tab:blimp}, the GPT2-Large and GPT2-Large-PPO models outperform the GPT-base variants on the BLiMP task with an average score of 73.9, excelling in many specific tasks. For example, GPT2-Large does well in tasks like Island Effects, NPI Licensing, and Subject-Verb Agreement, whereas GPT2-Large-PPO stands out in the  QA Congruence (tricky) task. The GPT2-Base and GPT2-Base-PPO models score lower with averages of 71.2 and 70.9, respectively, suggesting that model size (base versus large) plays a crucial role in determining performance.  However, for the BLiMP benchmark, PPO training has little impact on model performance. However, more experiments on different architecture could potentially point in a different direction.

\paragraph{Performance on SuperGLUE benchmarks.} In Table~\ref{tab:my-glue}, we report the performance of the models on the SuperGLUE benchmarks, which assess a range of language understanding abilities. The GPT2-Large-PPO model stands out with the highest average score of 66.8, underlining the potential for enhanced performance using larger models fine-tuned with PPO. Other models present comparable average scores across the SuperGLUE tasks. Compared to the Majority Label baseline, the GPT-2 models exhibit varied levels of performance enhancement across different tasks. Specifically, the GPT2-Base model outperforms the baseline in SST-2, QQP (F1), MNLI, MNLI-mm, QNLI, and BoolQ. Similarly, the GPT2-Base-PPO model surpasses the baseline in the same tasks: SST-2, QQP (F1), MNLI, MNLI-mm, QNLI, and BoolQ. The GPT2-Large model demonstrates superior performance over the baseline in SST-2, MRPC (F1), MNLI, MNLI-mm, QNLI, BoolQ, and WSC. While, the GPT2-Large-PPO model outperforms the majority baseline in all tasks except for CoLA and MultiRC, marking significant performance improvement in SST-2, MNLI-mm, and QNLI, with an increase of 34.1, 28.3, and 44.2 respectively.

The performance across various models and tasks exhibits considerable variability, showing that different models may excel in distinct language understanding domains.
The superior scores of the GPT2-Large-PPO model suggest that larger models fine-tuned with PPO could enhance performance, yet further examination reveals inconsistencies. Finally, we note that the PPO training only improves the performance of the GPT2-Large model, suggesting that PPO training may require a model with a minimum number of parameters to work in the limited data setting. However, more experiments are needed to confirm this finding.

\paragraph{Performance on MSGS benchmarks}

Table~\ref{tab:my-msgs} shows the results of testing GPT2 models of different sizes on the MSGS benchmark. These results help us understand how well the models use and generalize different language and surface features. Among the models, the GPT2-Base model outperforms other models with the highest average score of 83.0. This suggests that GPT2-Base, despite being a smaller model, has effectively learned to generalize across a range of linguistic and surface features. This might be due to the model's efficient use of its limited parameters. Instead of overfitting to less important details in the training data. 


\paragraph{Performance on Age-of-acquisition benchmarks}


According to \citet{portelance2023predicting}, a smaller mean absolute deviation (MAD) score indicates a better alignment between the model's predictions and the actual average age-of-acquisition (AoA) of words in children. Table~\ref{tab:my-age} shows similar MAD scores across all models for all word categories (Overall, Nouns, Predicates, and Function words). This suggests that all models exhibit similar levels of accuracy in predicting the AoA of words, and their word-learning sequences align closely with the natural language acquisition patterns observed in children.

\begin{table}[t]
\resizebox{\linewidth}{!}{%
\begin{tabular}{@{}lrrrr@{}}
\toprule
\textbf{Model} &
\textbf{Overall} &
\textbf{Nouns} &
\textbf{Predicates} &
\textbf{Function words} \\ \midrule
GPT2-Base      & 2.05 & 1.98 & 1.84 & 2.62 \\
GPT2-Base-PPO  & 2.06 & 1.99 & 1.83 & 2.66 \\ \midrule
GPT2-Large     & 2.05 & 1.98 & 1.83 & 2.63 \\
GPT2-Large-PPO & 2.05 & 1.98 & 1.82 & 2.63 \\ \bottomrule
\end{tabular}%
}
\caption{Performance on the Age-of-acquisition benchmarks. This table presents Mean Absolute Deviation (MAD) scores in months, comparing the actual average age-of-acquisition (AoA) of words by American English speaking children with the predicted AoA based on the model's average surprisal scores. A lower MAD score indicates a better fit between the actual and predicted AoA. }
\label{tab:my-age}
\end{table}

\begin{table}[t]
\resizebox{\linewidth}{!}{%
\begin{tabular}{@{}lrrrr@{}}
\toprule
\textbf{Model} &
 \textbf{Gram.} &
\textbf{Creat.} &
\textbf{Consist.} &
\textbf{PCoh} \\ \midrule
GPT2-Base      & 7.84 & 6.11 & 3.49 & 1.94 \\
GPT2-Base-PPO  & 6.82*** & 5.66** & 3.37 & 1.89 \\ \midrule
GPT2-Large     & 7.65 & 6.23 & 3.94 & 2.35 \\
GPT2-Large-PPO & 7.62 & 6.21 & 4.48** & 2.87* \\ \bottomrule 
\end{tabular}%
}
\caption{Performance on Human Evaluation. Gram. (Grammar), Creativity (Creat.), Consistency (Consist.), and Plot Coherence (PCoh). Significant
differences based on a paired t-test are marked with * (p-value < .1), ** (p-value < .05), and *** (p-value <.001).}
\label{tab:human-score}
\end{table}

\begin{table*}[t]
\centering
\resizebox{\textwidth}{!}{%
\begin{tabular}{@{}p{4cm}p{4cm}p{5cm}p{4cm}p{4cm}@{}}
\toprule
Prompt & gpt2-base & gpt2-base-ppo & gpt2-Large & gpt2-large-ppo \\ \midrule
Aquarium closed Yesterday I went to the aquarium. & It was on the way to the museum and there was a lot of fish in it. \ \  (10/7/7/7) & It was one of the most important things I've ever seen in my life. The first thing I want to tell you is that we are not going to be able to help each other. The first thing I want to tell you is that we are not going to be able to help each other. We are going to be able to help each other. (6/6/1/1)  & " It was full of fish, and there were so many different kinds of things to eat. "  (10/8/8/7)  & " It was full of fish. He looked at it and he said, ""I'm going to tell you a story."" (10/9/8/7) \\
Awkward I was driving into the McDonald's beside school. & I'm going to tell you something,"" he said. ""You're going to have to do it, don't you? " (9/7/4/1) & "I don't know what to do,"" he said. ""I'm not going to tell anyone about it."" " (9/7/4/1) & It was about two-and-a-half hours ahead of my time. It was then that I realized that I had to get back to work. (9/7/4/1) & "I didn't know how to drive, but I thought it was great fun. " (10/7/7/7)  \\
Tie Game The teams were tied and this was the last inning. & It was also the first time that Tie Game would be able to play the part of the game. (6/2/9/8) & " ""Tie Game"" was released on August 26, 2010. It is the second single from Tie Game." (8/7/8/1) & It's hard to imagine how easy it would have been for them to come up with us. (9/6/2/1) & " That's what I'm trying to figure out, so let's try it again. " (9/6/2/1)\\ \bottomrule
\end{tabular}}
\caption{Performance comparison of various models on grammar, creativity, consistency with the beginning of the story, and plot coherence. The scores in the parentheses represent the evaluations for Grammar, Creativity, Consistency, and Plot, respectively.}
\vspace{-1em}
\label{tab:human-eval}
\end{table*}

\paragraph{Performance on Human Evaluation.} In Table~\ref{tab:human-score}, we report the results of our human evaluation. The findings indicate that the GPT2-Base and GPT2-Large models exhibit comparable average grammar scores. However, the GPT2-Base-PPO model performs significantly worse (p-value < 0.001) than the GPT2-Base in grammar and creativity evaluations. The result is consistent with the BablyLM automated evaluation metrics, where the GPT-Base-PPO generally underperforms GPT-Base.  Table~\ref{tab:human-eval} shows several examples from our TinyStory analysis. Specifically, the GPT2-Base-PPO tends to generate repetitive and lengthy stories, likely contributing to its poorer grammar and creativity performance. Furthermore, when comparing the GPT2-Large and GPT2-Large-Base models in Table~\ref{tab:human-score}, their performance levels for Grammar and Creativity are similar, showing that PPO had minimal impact on the Large model for both metrics.

We also find significant differences in Consistency (Const.) and Plot Coherence (PCoh) between GPT-Large and GPT2-Large-PPO. Intuitively, these metrics evaluate generative models' capability in following the beginning of the story background rather than just content creation. Our findings indicate that the performance scores for GPT2-Base and GPT2-Base-PPO models are fairly similar, but both are lower than those of the GPT2-Large model variants. Again, this indicates that the large models outperform the smaller models, even though we trained on a relatively small dataset. Moreover, the GPT2-Large-PPO model significantly improves consistency and plot coherence scores compared to the standard GPT2-Large model. This suggests that large models (at least GPT2-Large in our case) can integrate the reward model to generate better outputs than the GPT2-base (smaller model).

We analyze the large model outputs in Table~\ref{tab:human-eval}. Specifically, in the second story from Table~\ref{tab:human-eval}, the beginning of the story is set as ``Awkward I was driving into the McDonald’s beside school.'' Distinct differences can be seen when comparing the narrative continuations generated by the GPT2-Large and GPT2-Large-PPO models. For example, the GPT2-Large model diverges from the initial context, transitioning abruptly from the act of driving into McDonald's to a sudden need to return to work. This abrupt shift disrupts the narrative flow and doesn't seamlessly connect with the story's beginning. On the other hand, the GPT2-Large-PPO model manages to retain focus on the primary activity of driving in its generated story. Although it introduces an inconsistency by stating the character doesn't know how to drive, it maintains the plot around the theme of a character recklessly driving without knowing how to do so. This suggests that the GPT2-Large-PPO model has a stronger adherence to the initial instructions and makes a better attempt at following them.

\paragraph{Summary of Findings and Limitations.} Overall, we found that the \textbf{GPT-2-Large generally works better than GPT-2-base \textit{with} and \textit{without} PPO}. Also, \textbf{PPO made significant improvements to the model's consistency and plot coherence on the storytelling task when used with the large model}. However, PPO generally hurts performance with the smaller GPT-2-Base model.

There were several limitations to our study. First, a major limitation of this work is the lack of comparison with architectures beyond GPT-2. Moreover, comparisons to even larger models should be made in the future. We were limited by the computational resources required for large-scale testing during the BabyLM shared task timeline. Next, we had a limited-size reward model dataset. Future work should explore the impact of reward model dataset size and variety. Future work should explore the impact of reward model dataset size and variety. Additionally, the study did not explore the hyperparameter tuning for the reward model and the loss function in depth. Exploring different settings for hyperparameters and examining alternative methods for reward training, such as varying the weighting of the loss terms, could yield different results and improve the model's performance in the storytelling task. Finally, we only had one annotator for the human evaluation and were limited in size. A more extensive human study could find more intricate differences between the models.

\section{Conclusion}

In this study, we investigated whether the small pretrained model, with its limited data size, can also benefit from RLHF, thus potentially improving its overall performance. We evaluate the two variants of the GPT-2 model: the GPT-2 Base model with 125M parameters and the larger GPT-2 Large model with 774M parameters. Both variants are pretrained on the 100M words BabyLM Challenge dataset. We then fine-tune both models using RLHF and evaluate their ability to acquire new linguistic patterns and storytelling ability, including generating coherent and creative English text while adhering to the story background. We observe that RLHF has a little or negative effect on the smaller model. However, a substantial increase in model parameters noticeably enhances the larger model's performance in storytelling tasks. In summary, our experiments shed light on the behavior of small language models fine-tuned using RLHF to perform storytelling tasks in a limited dataset setting.



\bibliography{anthology,custom}
\bibliographystyle{acl_natbib}





\label{sec:appendix}


\end{document}